\documentclass{llncs}
\usepackage{multicol}
\usepackage{graphicx}
\usepackage{xspace}
\usepackage{hyperref}
\usepackage{paralist}
\usepackage{amssymb,amsmath}
\usepackage{ifthen}
\usepackage{multirow}
\usepackage{array}
\usepackage{booktabs}
\usepackage{url}
\usepackage[obeyFinal]{todonotes} 
\usepackage{array}
\usepackage{amsmath}
\usepackage{multicol}
\usepackage{listings}
\usepackage{wrapfig}
\usepackage{totcount}
\usepackage{enumitem}
\usepackage{colortbl}
\usepackage{xcolor}
\usepackage{changepage}
\title{Predictive Process Monitoring Methods: \\Which One Suits Me Best?}
\author{Chiara Di Francescomarino\inst{1} \and Chiara Ghidini\inst{1} \and \\Fabrizio Maria Maggi\inst{2} \and Fredrik Milani\inst{2}}
\institute{FBK-IRST, Via Sommarive 18, 38050 Trento, Italy \\
{\fontsize{8}{10}\selectfont\email{ \{dfmchiara, ghidini\}@fbk.eu}
}
\and University of Tartu, Liivi 2, 50409 Tartu, Estonia \\
{\fontsize{8}{10}\selectfont\email{\{f.m.maggi, milani\}@ut.ee}}
}
\urldef{\mailsc}\path||
\newcommand{\keywords}[1]{\par\addvspace\baselineskip
\noindent\keywordname\enspace\ignorespaces#1}

\begin{document}

\maketitle
\setcounter{footnote}{0}
\begin{abstract}
Predictive process monitoring has recently gained traction in academia and is maturing also in companies. However, with the growing body of research, it might be daunting for companies to navigate in this domain in order to find, provided certain data, what can be predicted and what methods to use. The main objective of this paper is developing a value-driven framework for classifying existing work on predictive process monitoring. This objective is achieved by systematically identifying, categorizing, and analyzing existing approaches for predictive process monitoring. The review is then used to develop a value-driven framework that can support organizations to navigate in the predictive process monitoring field and help them to find value and exploit the opportunities enabled by these analysis techniques. 
\keywords{Predictive Process Monitoring, Process Mining, Systematic Literature Review, Value-Driven Framework}
\end{abstract}

\section{Introduction}

Utilizing sets of data has become the modus operandi of business development. However, until recently, in the day to day operations, companies have relied on limited availability of data processing capacities to estimate simple business parameters. Therefore, oftentimes, these analyses have been ``guesstimates'' rather than ``estimates''. Nowadays, companies are replacing ``gut feeling'' with data-driven results and with the improvement of data analysis, automated methods can extend business insights far further than what the human mind could ever hope to achieve. 

Predictive process monitoring represents one of these enablers of data-driven insights. Since nowadays there is a large availability of data logged by information systems supporting the day to day operations of companies, and mature and fast approaches are rapidly growing in the predictive process monitoring field, these opportunities are becoming more and more handy. 


Nevertheless, although selected companies have integrated such predictive methods in their business processes, the potential for greater impact is still very real. However, due to the large availability of techniques, it might be daunting for companies to navigate in this domain. In the quest to find, provided certain data, what can be predicted and what methods to use, companies can easily get lost. As such, this paper has the objective, to develop, based on review of existing research, a value-driven framework for classifying existing work on predictive process monitoring. 

This objective is achieved by a systematic literature review to identify the state of the art of predictive process monitoring. The aim is to identify, categorize, and analyze existing methods. The review is then used to develop the framework that supports organizations to identify the methods that fit their needs (in the large body of research available) and grab the opportunities provided by these analysis techniques.

The remainder of the paper is organized as follows. Section 2 describes the background while Section 3 presents the literature review protocol and the results from it. Section 4 provides a classification of the identified methods. In Section 5, the framework is presented and discussed. Finally, Section 6 concludes the paper.

\section{Background}
\label{sec:background} 
\subsection{Process Mining}

The starting point for process mining is an event log \cite{DBLP:books/sp/Aalst16}. Each event in a log refers to an \emph{activity} (i.e., a well-defined step in some process) and is related to a particular \emph{case} (i.e., a \emph{process instance}). The events belonging to a case are \emph{ordered} and can be seen as one ``run'' of the process (often referred to as a \emph{trace} of events). Event logs may store additional information about events such as the \emph{resource} (i.e., person or device) executing or initiating the activity, the \emph{timestamp} of the event, or \emph{data elements} recorded with the event. Data elements can be event attributes, i.e., data produced by the activities of a business process and case attributes, namely data that are associated to a whole process instance. 
Predictive process monitoring belongs to the family of process mining techniques since the information used to make predictions is retrieved from logs of historical traces.



\subsection{Predictive Process Monitoring}
\label{sec:predictive}
The execution of business processes is generally subject to internal policies, norms, best practices, regulations, and laws.
%
%
For this reason, compliance monitoring is an everyday imperative in many organizations. Accordingly, a range of research proposals have addressed the problem of assessing whether a process execution complies with a set of requirements \cite{DBLP:journals/is/LyMMRA15}. 
However, these monitoring approaches are \emph{reactive}, in that they allow users to identify a violation only \emph{after it has occurred} rather than supporting them in \emph{preventing} such violations in the first place.
%
%
%


Based on an analysis of historical execution traces, \textit{predictive process monitoring} methods \cite{Maggi:CAiSE2014} continuously provide the user with predictions about the future of a given ongoing process execution. The forward-looking nature of predictive monitoring provides organizations with new frontiers in the context of compliance monitoring. Indeed, knowing in advance about violations, deviances and delays in a process execution would allow them to take preventive measures (e.g., reallocating resources) in order to avoid the occurrence of those situations and, as often happens, to avoid money loss.

\section{Systematic Review Protocol}
\label{sec:protocol}

The systematic review protocol of our literature review specifies the research questions, the search protocol, and the selection criteria predominantly following the guidelines provided by Kitchenham et al.\ \cite{kitchenham2004procedures}. The work was divided into two phases. In the first phase, two researchers designed the review protocol. Here, the research questions were formulated, electronic databases identified, inclusion and exclusion criteria defined, and data extraction strategy formulated. The second phase was conducted by two other researchers who reviewed the protocol, ran the searches, filtered the list of papers, produced the final list of papers and extracted the data.

The objective of the systematic literature review is to create a value-driven framework for predictive process monitoring. To this end, it examines the main research question (\textbf{RQ}): ``\textit{what is the body of relevant academic publications within the field of predictive process monitoring?}''. The main research question is decomposed into four sub-categories. These sub-research questions investigate, with regards to business processes, the following aspects of predictive process monitoring: type of prediction, input data required, type of algorithm employed, validation method, and tool support.

As such, the main research question is decomposed in the following four sub-questions.
The first research question seeks to identify the different aspects of business processes that can be predicted by means of predictive process monitoring techniques. As such, the first sub-research question is formulated as \textbf{RQ1}: ``\textit{what aspects of business processes does predictive process monitoring predict?}''. The next two research questions concern the algorithms employed in predictive process monitoring techniques. The second is \textbf{RQ2}: ``\textit{what input data do predictive process monitoring algorithms require?}''. The third research questions is \textbf{RQ3}: ``\textit{what are the main families of algorithms used in predictive process monitoring techniques?}''. Finally, the last research question focuses on tools for predictive process monitoring i.e., \textbf{RQ4}: ``\textit{what are the tools that support predictive process monitoring?}''.

The first step was to define the search string. We followed the guidance given by \cite{kitchenham2004procedures}, which resulted in using the keywords ``predictive'', ``prediction'', ``business process'', and ``process mining''. The keywords ``predictive'' and ``prediction'' were derived from the research questions. However, the literature on predictive analysis is vast and encompasses areas outside of the domain of business processes. As such, we added the keyword ``business process''. Finally, predictive process monitoring concerns ``process mining'' and, therefore, this keyword was added. To retrieve as many results as possible, we intentionally left out additional keywords such as ``monitoring'', ``technique'', and ``algorithm'' so not to limit the search. The keywords were used to formulate the following boolean search string; (``predictive'' OR ``prediction'') AND (``business process'' OR ``process mining'').

Following the definition of the search string, the electronic libraries were chosen. The databases selected were Scopus, SpringerLink, IEEE Xplore, Science Direct, ACM Digital Library, and Web of Science.  These were selected as they cover scientific publications within the field of computer science.

The results were exported into an excel sheet for processing. The first filtering was for duplicates. Duplicate studies are those that appeared in more than one electronic database with identical title by the same author(s) \cite{kofod2012structured}. Likewise, papers not written in English or inaccessible were excluded. Having performed this filtering, 779 papers were identified. Next, the list was filtered based on title of the study. Studies clearly out of scope were removed. After this filter, 186 papers remained. The filtering continued with regards to venue, length and citations. Papers with less than 6 pages and published in workshops were excluded as their contribution is less mature as compared to full paper published in conferences and journals. At this stage, 162 papers remained. The next filtering was based on the number of citations. Papers published in 2016 and later were all included, even if they had no citations. However, for papers published before 2016, the requirement was to have at least 10 citations. This criterion was set to filter out the least relevant papers. Having applied this filter, 92 papers remained. We then filtered the papers by looking at the abstracts to further assess their relevancy. Following this step, 73 papers remained. Finally, we used the inclusion criterion ``\textit{does the study propose a novel algorithm or technique for predictive process monitoring?}''. We found 51 papers that complied with this criterion.

From the final list of 51 papers, we extracted standard meta-data (title, authors, year of publication, number of citations, and type of publication). For each paper, the type of prediction considered was extracted in accordance with \textbf{RQ1}. Some methods can predict several aspects of a business process. In such cases, all aspects were recorded. Secondly, information about the input data was extracted. Process mining requires that logs contain at least a unique case id, activity names and timestamps. However, if a method requires additional data for analysis, this is extracted and noted (\textbf{RQ2}). Thirdly, we examined the type of algorithm. The underlying family of each identified predictive process monitoring method was extracted to address \textbf{RQ3}. 
We also extracted information about the validation of each method. The quality of a method depends indeed on its validation. As such, data about validation was extracted. In addition, if validated, data about if the log was synthetic or from real life (including industry domain) was extracted as well. Data about tool support was also extracted. This information considers if there is a plug-in or a standalone application for the proposed predictive process monitoring technique. This data relates to \textbf{RQ4}.
In addition to the above data, the type of output, the metrics used for evaluating the algorithms and if the datasets used were public or not was annotated.\footnote{The data extracted was entered into an excel sheet and is available for download at \url{https://docs.google.com/spreadsheets/d/1l1enKhKWx_3KqtnUgggrPl1aoJMhvmy9TF9jAM3snas/edit\#gid=959800788}.}


\section{Predictive Process Monitoring Dimensions}
\label{sec:classification} 


Typically, the literature on predictive process monitoring can be classified along three main dimensions:
\begin{small}
\begin{itemize}
\item [\textbullet] type of prediction (provided as output);
\item [\textbullet] type of algorithm (used to predict);
\end{itemize}
\end{small}

Concerning the type of prediction, we can classify the existing prediction types into three main macro-categories:
\begin{itemize}
\small
\item [\textbullet] predictions related to numeric or continuous measures of interest (\emph{numeric predictions}). Typical examples in this configuration are predictions related to the remaining time of an ongoing execution and predictions related to the duration of an ongoing case or to its cost;
\item [\textbullet] predictions related to categorical or boolean outcomes (\emph{categorical predictions}). Typical examples in this configuration are predictions related to the class of risk of a given execution or to the outcome of a predicate along the lifecycle of a case;
\item [\textbullet] predictions related to sequences of future activities
 (\emph{activity sequence predictions}). Typical examples of predictions falling under this category refer to the prediction of the sequence of the future activities (and of their payload) of a process case upon its completion.
\end{itemize}

Predictive process monitoring approaches are usually characterized by two phases. In a first phase, the \emph{training or learning} phase, one or more prediction models are built or enriched by leveraging the information contained in the execution log. In the second phase, the \emph{runtime or prediction phase}, the prediction model(s) is(are) exploited in order to get predictions related to one or more ongoing execution traces.
We can identify two main groups of approaches dealing with the prediction problem. They differ significantly on the way the predictive model is built.
\begin{itemize}
\small
\item[\textbullet] approaches relying on an explicit model, e.g., annotated transition systems. The explicit model can either be discovered from the event log or enriched with the information the log contains, if it is already available.
\item[\textbullet] approaches leveraging machine learning and statistical techniques, e.g., classification and regression models, or neural networks. These approaches only rely on (implicit) predictive models built by encoding event log information in terms of features to be used as input for machine/deep learning techniques.
\end{itemize}



In the following sections, some of the research works characterizing each of the three prediction type macro-categories, i.e., numeric, categorical and activity sequence predictions, are presented. They are further classified based on the specific output type and approach.

\subsection{Numeric Prediction}
We can roughly classify the works dealing with numeric predictions in two groups, based on the specific type of predictions returned as output:
\begin{small}
\begin{itemize}
\item[\textbullet] time predictions;
\item[\textbullet] cost predictions;
\end{itemize}
\end{small}

\textbf{Time predictions.}
The group of works focusing on the time perspective is a rich group.
Several works, in this context, rely on explicit models. In~\cite{Aalstetal:2011}, the authors present a set of approaches in which transition systems, which are built based on a given abstraction of the events in the event log, are annotated with time information extracted from the logs. Specifically, information about elapsed, sojourn, and remaining time is reported for each state of the transition system. The information is then used for making predictions on the completion time of an ongoing trace. Further extensions of the annotated transition system are proposed in~\cite{Polato:2014,Polatoetal:2018}, where the authors apply machine learning techniques to the annotated transition system. In detail, in~\cite{Polato:2014}, the transition systems are annotated with machine learning models such as Na\"{i}ve Bayes and Support Vector Regression models. In~\cite{Polatoetal:2018}, instead, the authors present, besides two completely new approaches based on Support Vector Regression, a refinement of the work in~\cite{Polato:2014} that takes into account also data. Moreover, the authors evaluate the three proposed approaches both on stationary and non-stationary (i.e., characterized by evolving conditions) processes. Other extensions of the approach presented in~\cite{Aalstetal:2011} that also aim at predicting the remaining time of an ongoing trace, are the works presented in~\cite{Folino,Folino2013}. In these works the annotated transition system is combined with a context-driven predictive clustering approach. The idea behind predictive clustering is that different scenarios can be characterized by different predictors. Moreover, contextual information is exploited in order to make predictions, together with control-flow (\cite{Folino}) or control-flow and resources (\cite{Folino2013}).

Another approach based on the extraction of (explicit) models (\emph{sequence trees}) is presented in~\cite{Ceci2014} to predict the completion time and the next future activity of the current ongoing case. Similarly to the predictive clustering approach, the sequence tree model allows for clustering traces with a similar sequence of activities (control-flow) and to build a predictor model for each node of the sequence tree by leveraging data payolad information.
In~\cite{DBLP:conf/icsoc/Rogge-SoltiW13}, the authors use generally distributed transitions stochastic Petri nets (GDT-SPN) to predict the remaining execution time of a process instance. In detail, the approach takes as input a stochastic process model, which can be known in advance or inferred from historical data, an on-going trace, and some other information as the current time in order to make predictions on the remaining time.
In~\cite{Rogge-Soltietal:2015}, the authors also exploit the elapsed time since the last event in order to make more accurate predictions on the remaining time and estimating the probability to miss a deadline.

Differently from the previous approaches, in~\cite{vanDongenetal2008}, the authors only rely on the event log in order to make predictions. In detail they
 develop an approach for predicting the remaining cycle time of a case by using non-parametric regression and leveraging activity duration and occurrence as well as other case-related data. In~\cite{Bevacqua2013} and~\cite{Cesario2016}, the contextual clustering-based approach presented in~\cite{Folino,Folino2013} is updated in order to address the limitation of transition system based approaches requiring the analyst to choose the log abstraction functions,
by replacing the transition system predictor with standard regression algorithms. Moreover, in~\cite{Cesario2016}
the clustering component of the approach is further improved in order to address scalability and accuracy issues.
In~\cite{Pandey:2011} Hidden Markov Models (HMM) are used for making predictions on the remaining time. A comparative evaluation shows that HMM provides  more accurate results than annotated transition systems and regression models. 
In~\cite{Senderovich2017} \emph{inter-case feature predictions} are introduced for predicting the completion time of an on-going trace. The proposed approaches leverage not only the information related to the ongoing case, but also the status of other (concurrent) cases (e.g., the number of concurrent cases) in order to make predictions. The proposed encodings demonstrated an improvement of the results when applied to two real-life case studies.

A prediction type which is very close to time but slightly different is the prediction of the delay of an ongoing case.
In~\cite{Senderovichetal15}, queuing theory
 is used to predict possible online delays in business process executions. The authors propose approaches that either enhance traditional approaches based on transition systems, as the one in~\cite{Aalstetal:2011}, to take queueing effects into account, or leverage properties of queue models.

\textbf{Cost predictions.}
A second group of works focuses on cost predictions. Also in this group, we can find works explicitly relying on models as the work in~\cite{Tuetal:2016}. In such a work, cost predictions are provided by leveraging a process model-enhanced cost (i.e., a frequent-sequence graph enhanced with cost) taking into account information about production, volume and time.

\subsection{Categorical Predictions}
The second family of prediction approaches predicts categorical values. In this settings, two main specific types of predictions can be identified:
\begin{small}
\begin{itemize}
\item[\textbullet] risk predictions;
\item[\textbullet] categorical outcome predictions.
\end{itemize}
\end{small}

\textbf{Risk predictions.}
A first large group of works falling under the umbrella of outcome-oriented predictions, deals with the prediction of risks.

Also in this case an important difference among state-of-the-art approaches is the existence of an explicit model guiding the prediction.
For example, in~\cite{DBLP:conf/caise/ConfortiLRA13}, the authors present a technique for reducing process risks. The idea is supporting process participants in making risk-informed decisions, by providing them with predictions related to process executions. Decision trees are generated from logs of past process executions, by taking into account information related to data, resources and execution frequencies provided as input with the process net. The decision trees are then traversed and predictions about risks returned to the users. In~\cite{Confortietal2015} and~\cite{Confortietal:2016}, two extensions of the work in~\cite{DBLP:conf/caise/ConfortiLRA13} are presented. In detail, in the former, i.e.,~\cite{Confortietal2015}, the framework for risk-informed decisions is extended to scenarios in which multiple process instances run concurrently. Specifically, in order to deal with the risks related to different instances of a process, a technique that uses integer linear programming is exploited to compute the optimal assignment of resources to tasks to be performed. In the latter, i.e.,~\cite{Confortietal:2016}, the work in~\cite{DBLP:conf/caise/ConfortiLRA13} is extended so that the process executions are no more considered in isolation but, rather, the information about risks is automatically propagated to similar running instances of the same process in real-time in order to provide early runtime predictions.

In~\cite{Metzgeretal2015}, three different approaches for the prediction of process instance constraint violation are investigated: machine learning, constraint satisfaction and QoS aggregation. The authors, beyond demonstrating that all the three approaches achieve good results, identify some differences and propose their combination. Results on a real case study show that combining these techniques actually allows for improving the prediction accuracy.

Other works, devoted to risk prediction, do not take into account explicit models. For instance in~\cite{Pika}, the authors make predictions about time-related process risks by identifying and leveraging process risk indicators (e.g., abnormal activity execution time or multiple activity repetition) by applying statistical methods to event logs. The indicators are then combined by means of a prediction function, which allows for highlighting the possibility of transgressing deadlines. In~\cite{Pikaetal2013b}, the authors extend their previous work by introducing a method for configuring the process risk indicators. The method learns from the outcomes of completed cases the most suitable thresholds for the process risk indicators, thus taking into account the characteristics of the specific process and hence improving the accuracy.

\textbf{Categorical outcome predictions.}
A second group of predictions relates to the fulfillment of predicates. Almost all works falling under this category do not rely on any explicit model. For example, in~\cite{Maggi:CAiSE2014} a framework for predicting the fulfillment (or the violation) of a predicate in an ongoing execution, is introduced. Such a framework makes predictions by leveraging: (i) the sequence of events already performed in the case; and (ii) the data payload of the last activity of the ongoing case. The framework is able to provide accurate results, although it demands for a high runtime overhead. In order to overcome such a limitation, the framework has been enhanced in~\cite{Di-Francescomarino:2016aa} by introducing a clustering preprocessing step in which cases sharing a similar behaviour are clustered together. A predictive model - a classifier - for each cluster is then trained with the data payload of the traces in the cluster. 
 In~\cite{DiFrancescomarinoetal:2016}, the framework is enhanced in order to support users by providing them with a tool for the selection of the techniques and the hyperparameters that best suit their datasets and needs. 
 
In~\cite{Leontjeva2015}, the authors consider traces as complex symbolic sequences, that is, sequences of activities each carrying a data payload consisting of attribute-value pairs. By starting from this assumption, the authors focus on the comparison of different feature encoding approaches, ranging from traditional ones, such as counting the occurrences of activities and data attributes in each trace, up to more complex ones, relying on Hidden Markov Models (HMM).  In~\cite{Verenich2016}, the approach in~\cite{Leontjeva2015} is enhanced with clustering, by proposing a two-phase approach. In the first phase, prefixes of historical cases are encoded as complex symbolic sequences and clustered. In the second phase a classifier is built for each of the clusters. At runtime, (i) the cluster closest to the current ongoing process execution is identified; and (ii) the corresponding classifier  is used for predicting whether the process instance outcome (e.g., whether the case is normal or deviant).

In~\cite{DBLP:conf/bpm/TeinemaaDMF16}, in order to improve prediction accuracy, unstructured (textual) information, contained in text messages exchanged during process executions, is also leveraged, together with control and data flow information. Specifically, different combinations of text mining (bag-of-n-grams, Latent Dirichlet Allocation and Paragraph Vector) and classification (random forest and logistic regression) techniques  have been proposed and exercised. In~\cite{Marquez-Chamorro2017}, an approach based on evolutionary algorithms is presented. The approach, which uses information related to a window of events in the event log, is based on the definition of process indicators (e.g., whether a process instance is completed in time, whether it is reopened). At training time, the process indicators are computed, the training event log encoded and the evolutionary algorithms applied for the generation of a predictive model composed of a set of decision rules. At runtime, the current trace prefix is matched against the decision rules in order to predict the correct class for the ongoing running instance.

\subsection{Activity Sequence Predictions}

A third more recent family of works deals with predicting the sequence of 
the future activities and their payload given the activities observed so far, as in~\cite{
Polatoetal:2018,Taxetal:2017,evermann,DiFrancescomarinoetal:2017}.
In~\cite{Polatoetal:2018}, the authors propose an approach for predicting the sequence of future activities of a running case by relying on an annotated data-aware transition system, obtained as a refinement of the annotated transition system
 proposed in~\cite{Aalstetal:2011}.

Other approaches, e.g., \cite{evermann,Evermann2017,Taxetal:2017}, make use of RNNs with LSTM cells.
In particular,
in~\cite{evermann,Evermann2017} an RNN with two hidden layers trained with back propagation is presented,  
while
in~\cite{Taxetal:2017} an LSTM and an encoding based on activities and timestamps is leveraged to provide predictions on the next activities and their timestamps. Finally, the work in~\cite{DiFrancescomarinoetal:2017} investigates how to take advantage of possibly existing a-priori knowledge for making predictions on the sequence of future activities. To this aim, an LSTM approach is equipped with the capability of taking into account also some a-priori sure knowledge.

\section{Value-driven Framework for Selecting Predictive Monitoring Algorithm}
\label{sec:evaluation}

The literature review reveals that the number of publications on predictive process monitoring has significantly increased in the last few years. 
In parallel, industry, in order to replace ``gut-driven'' estimates, is increasingly seeking the benefits that predictive process monitoring, by enabling data driven decisions, is able to provide. However, there are no ready-made predictive process monitoring solutions that can be plugged in and used. Companies can either develop their own algorithms or turn to available resources. The second option is probably the more efficient as companies can tap into accumulated knowledge and experience already made and documented. As such, companies can adopt an existing academic solution and customize it to their own setting. However, algorithms of varying robustness, measuring different aspects, and requiring various sets of data to function, pose an obstacle where companies might not see the forest for the trees. In light of these obstacles, our value-driven framework serves to support companies in finding the best suited academic research for predictive process monitoring. The framework considers prediction type, input data required, tool support, validity of the algorithm, family of algorithm, and finally provides a reference to the specific work in the literature. 

\setlength{\aboverulesep}{2pt}
\setlength{\belowrulesep}{2pt}

\newcommand{\none}{\cellcolor{gray!90}}

\begin{table} 
\begin{adjustwidth}{-2cm}{0cm}
	\centering
	\scalebox{0.6}{
		\begin{tabular}{|l| l|>{\hspace{0.2em}} l>{\hspace{0.2em}} l>{\hspace{0.2em}} l>{\hspace{0.2em}} l>{\hspace{0.2em}} l>{\hspace{0.2em}} l>{\hspace{0.2em}} l>{\hspace{0.2em}} l>{\hspace{0.2em}} l|}
		\toprule
		\textbf{Pred.} & \textbf{Det.\ Pred.} & \multirow{2}{*}{\textbf{Input 1}} & \multirow{2}{*}{\textbf{Input 2}} & \multirow{2}{*}{\textbf{Input 3}} & \multirow{2}{*}{\textbf{Tool}} & \multirow{2}{*}{\textbf{Domain}} & \textbf{Family of} & \textbf{Family of } & \textbf{Family of} & \multirow{2}{*}{\textbf{Refer.}} \\ 
		\textbf{type} & \textbf{type} & & & & & & \textbf{algorithm 1} & \textbf{algorithm 2} & \textbf{algorithm 3} & \\
		\toprule
		\multirow{23}{*}{\rotatebox[origin=c]{90}{\textbf{time}}} & maint. time & \multirow{7}{*}{event log (with timestamps)} & \none & \none & \multirow{7}{*}{N} & automotive &	time series &	probabilistic model & \none &	\cite{Ruschel2017} \\ \cline{2-2}\cline{7-11}		

		& activity & & \none & \none & & financial	& \multirow{2}{*}{queueing theory}	& \multirow{2}{*}{transition system} & \none	& \multirow{2}{*}{\cite{Senderovichetal2014,Senderovichetal15}} \\ 
		& delays & & \none & \none & & telecomm. & & & \none & \\ \cline{2-2}\cline{7-11}			
		& \multirow{20}{*}{rem. time} & & \none & \none & & telecomm. & stat. analysis & \none & \none & \cite{Boltetal2014} \\ \cline {7-11}	
		& & & \none & \none & & financial	& \multirow{2}{*}{regression} & \multirow{2}{*}{classification}	& \none & \multirow{2}{*}{\cite{Verenicheetal2017}}\\
		& & & \none & \none & & customer supp. & & & \none & \\	\cline {6-11}	
		& & & \none & \none & \multirow{2}{*}{ProM plugin} & public admin. & transition system & \none & \none & \cite{Aalst2010:Beyond,Aalstetal:2011}\\ \cline {7-11}
		& & & \none & \none & & customer supp. & stochastic Petri net & \none & \none & \cite{Rogge-Soltietal2015} \\ \cline {3-11}
		& & & \none & \none & Y but unavail. & unspecified & classification & time series & \none & \cite{Castellanos2005}\\ \cline {5-11}
		& & & \none & \none & \multirow{3}{*}{Y} & financial & \multirow{2}{*}{regression} & \multirow{2}{*}{classification} & \none & \multirow{2}{*}{\cite{Verenichetal2016}}\\ 		
		& & & \none & \none & & public admin. & & & \none & \\ \cline {7-11}			
		& & & \none & \none & & healthcare & stochastic Petri net & \none & \none & \cite{Senderovichetal2016}\\ \cline {5-11}			
		& & & \none & \none & \multirow{7}{*}{ProM plugin} & public admin. & regression & \none & \none & \cite{vanDongenetal2008}\\ \cline {7-11}		
		& & & \none & \none & & public admin. & transition system & regression & classification & \cite{Polato:2014}\\ \cline {7-11}
		& & event log (with timestamps) & \none & \none & & customer supp. &\multirow{3}{*}{transition system} & \multirow{3}{*}{regression} & \none & \multirow{3}{*}{\cite{Polatoetal:2018}}\\ 
		& & with data & \none & \none & & public admin. &  &  & \none & \\ 
		& & & \none & \none & & financial &  &  & \none & \\ \cline {7-11}
		& & & \none & \none & & financial & \multirow{2}{*}{stochastic Petri net} & \none & \none & \multirow{2}{*}{\cite{DBLP:conf/icsoc/Rogge-SoltiW13,Rogge-Soltietal:2015}} \\ 
		& & & \none & \none & & logistics &  & \none & \none & \\ \cline {5-11}
		& & & \multirow{2}{*}{inter-case metrics} & \none & \multirow{2}{*}{Y} & healthcare & \multirow{2}{*}{regression} & \none & \none & \multirow{2}{*}{\cite{Senderovich2017}}\\ 
		& & & & \none & & manufacturing &  & \none & \none & \\ \cline {3-11}
		& & event log (with timestamps)  & \none & \none & Y but unavail. & logistics & clustering & regression & \none & \cite{Folinoetal:2014}\\ \cline{5-11}
		& & with data & \none & \none & ProM plugin & logistics & clustering & transition system & \none & \cite{Folino}\\ \cline {5-11}
		& & and contextual information & labelling funct. & proc. model & ProM plugin & no validation & classification & \none & \none & \cite{deleonietal:2014}\\ \hline 
		
		\multirow{40}{*}{\rotatebox[origin=c]{90}{\textbf{categorical outcome}}} & \multirow{30}{*}{outcome} & act. durations and routing probab. & threshold(s) & proc. model & N & synthetic &	simulation &	stat. analysis & \none &	\cite{Sietal:2016} \\ \cline {3-11}
		& & \multirow{2}{*}{event log} & \multirow{2}{*}{labelling funct.} & \none & \multirow{2}{*}{Y implem.} & financial &	\multirow{2}{*}{prob. automata} &	\none & \none &	\multirow{2}{*}{\cite{Breukeretal:2016}} \\
		& & & & \none & & automotive &	&	\none & \none &	\\ \cline{3-11}
		& & & \none & \none & \multirow{4}{*}{N} & logistics &	neural network & constraint-sat. & QoS aggregation &	\cite{Metzgeretal2015} \\ \cline{7-11}
		& & & \multirow{13}{*}{labelling funct.} & \none & & healthcare &	probab. automata & classification &	\none &	\cite{Leontjeva2015}\\ \cline{7-11}
		& & & & \none & & logistics &	classification & &	\none &	\cite{Cabanillas2014}\\ \cline{7-11}		
		& & & & \none & & synthetic &	classification & &	\none &	\cite{Kang:2012}\\ \cline{5-11}		
		& & & & \none & \multirow{3}{*}{Y but unavail.} & synthetic &	probab. automata & \none &	\none &	\cite{Lakshmanan:2015}\\ \cline{7-11}
		& & event log (with timestamps) & & \none & & synthetic &	classification & neural network &	\none &	\cite{Maisenbacheretal:2017}\\ \cline{7-11}		
		& & with data & & \none & & healthcare &	clustering & classification &	\none &	\cite{Di-Francescomarino:2016aa}\\ \cline{5-11}		
		& & & & \none & \multirow{4}{*}{Y} & no valid. &	stat. analysis & \none &	\none &	\cite{Feldman:2013}\\ \cline{7-11}
		& & & & \none & & logistics &	stat. analysis & \none &	\none &	\cite{Metzgeretal12}\\ \cline{7-11}		
		& & & & \none & & financial &	\multirow{2}{*}{classification} & \none &	\none &	\multirow{2}{*}{\cite{Verenichetal2016}}\\ 		
		& & & & \none & & public admin. & & \none &	\none &	\\ \cline{5-11}		
		& & & & \none & \multirow{2}{*}{ProM pl.} & healthcare &	classification & \none &	\none &	\cite{Maggi:CAiSE2014}\\ 
		& & & & \none & & healthcare &	clustering & classification &	\none &	\cite{DiFrancescomarinoetal:2016}\\ \cline{5-11}		
		& & & & \none & ProM and & automotive &	\multirow{2}{*}{evol. algorithm} & \none &	\none &	\multirow{2}{*}{\cite{Marquez-Chamorro2017}}\\ 
		& & & & \none & Camunda pl. & healthcare & & \none &	\none &	\\ \cline{4-11}	
		& & & threshold(s) & \none & Y but unavail. & unspecified & classification & \none &	\none &	\cite{Castellanos2005}\\ \cline{3-11}	
		& & & \none & \none & Y & domotic &	stat. analysis & \none &	\none &	\cite{Ferillietal:2016}\\ \cline{5-11}	
		& & & \multirow{2}{*}{labelling funct.} & \none & Y but unavail. & healthcare & clustering & classification &	\none & \cite{Cuzzocreaetal:2016}	\\ 	\cline{5-11}
		& & event log (with timestamps) & & proc. model & ProM plugin & no validation & classification & \none &	\none & \cite{deleonietal:2014}	\\ \cline{4-11}	
		& & with data & threshold(s) & \none & ProM plugin & logistics & clustering & transition system &	\none & \cite{Folino}	\\ 	\cline{4-11}
		& & and contextual information & \multirow{2}{*}{clusters of behav.} & \none & \multirow{2}{*}{N} & logistics & \multirow{2}{*}{clustering} & \multirow{2}{*}{classification} &	\none & \multirow{2}{*}{\cite{Folinoetal:2011}}	\\ 
		& & & & \none & & manufacturing & & &	\none &	\\ \cline{3-11}	
		& & event log (with timestamps) & \multirow{3}{*}{labelling funct.} & \none & \multirow{3}{*}{N} & \multirow{3}{*}{financial} & \multirow{3}{*}{classification} & \multirow{3}{*}{text mining} &\none & \multirow{3}{*}{\cite{DBLP:conf/bpm/TeinemaaDMF16}}	\\ 
	  & & with data & & \none & &  &  &	& \none & \\ 
		& & and unstructured text & & \none & & & & & \none & \\ \cline{2-11}	
		& \multirow{5}{*}{next activity} & event log (with timestamps) & \none & \none & \multirow{2}{*}{Y} & \multirow{2}{*}{domotic} &  \multirow{2}{*}{stat. analysis} & \none & \none &  \multirow{2}{*}{\cite{Ferillietal:2017}} \\
		& & with data & \none & \none& & & & \none & \none & \\ \cline{3-11}
		& & event log (with timestamps) & \none & \none & {Y} & {domotic} &  {stat. analysis} & \none & \none &  \cite{Ferillietal:2016} \\  \cline{5-11}
		& & with data & \multirow{2}{*}{labelling funct.} & \multirow{2}{*}{proc. model} & \multirow{2}{*}{ProM plugin} & \multirow{2}{*}{no valid.} & \multirow{2}{*}{classification} & \none & \none & \multirow{2}{*}{\cite{deleonietal:2014}}\\
		& & and contextual information & & & & & & \none & \none & \\ \cline{2-11}		
		& last value & event log (with timestamps) &  \multirow{3}{*}{labelling funct.} & \multirow{3}{*}{proc. model} & \multirow{3}{*}{ProM plugin} & \multirow{3}{*}{no valid.} & \multirow{3}{*}{classification} & \none & \none & \multirow{3}{*}{\cite{deleonietal:2014}}\\
		& of an & with data & & & & & & \none & \none & \\ 
		& attribute & and contextual information & & & & & & \none & \none & \\
		\bottomrule
		\end{tabular}
		}
		\vspace{0.5cm}
	\caption{Predictive process monitoring framework: \texttt{time} and \texttt{categorical outcome} predictions}
	\label{tab:framework}
	\end{adjustwidth}
\end{table}

\begin{table}
\begin{adjustwidth}{-2cm}{0cm} 
	\centering
	\scalebox{0.6}{
		\begin{tabular}{|l| l|>{\hspace{0.2em}} l>{\hspace{0.2em}} l>{\hspace{0.2em}} l>{\hspace{0.2em}} l>{\hspace{0.2em}} l>{\hspace{0.2em}} l>{\hspace{0.2em}} l>{\hspace{0.2em}} l>{\hspace{0.2em}} l|}
		\toprule
		\textbf{Pred.} & \textbf{Det.\ Pred.} & \multirow{2}{*}{\textbf{Input 1}} & \multirow{2}{*}{\textbf{Input 2}} & \multirow{2}{*}{\textbf{Input 3}} & \multirow{2}{*}{\textbf{Tool}} & \multirow{2}{*}{\textbf{Domain}} & \textbf{Family of} & \textbf{Family of } & \textbf{Family of} & \multirow{2}{*}{\textbf{Refer.}} \\ 
		\textbf{type} & \textbf{type} & & & & & & \textbf{algorithm 1} & \textbf{algorithm 2} & \textbf{algorithm 3} & \\
		\toprule
		\multirow{21}{*}{\rotatebox[origin=c]{90}{\textbf{sequence of outcomes/values}}} &  &  \multirow{10}{*}{event log (with timestamps)} & \none & \none & \multirow{3}{*}{N} & customer supp. &\multirow{3}{*}{neural network}& \none & \none &\multirow{3}{*}{\cite{Taxetal:2017}} \\
	  & & & \none & \none & & financial & & \none &	\none &	\\ 
		& & & \none & \none & & public admin. & & \none &	\none &	\\ \cline{4-11}	
		& & & \none & \none & \multirow{3}{*}{Y but unavail.} & financial &\multirow{3}{*}{neural network}& \none & \none &\multirow{3}{*}{\cite{Mehdiyevetal:2017}} \\
	  & & & \none & \none & & automotive & & \none &	\none &	\\ 
		& & & \none & \none & & customer supp. & & \none &	\none &	\\ \cline{4-11}	
		& & & \none & \none & \multirow{2}{*}{Y} & financial &\multirow{2}{*}{neural network}& \none & \none &\multirow{2}{*}{\cite{Evermannetal:2017}} \\
		& & & \none & \none & & automotive & & \none &	\none &	\\ \cline{4-11}
		& & & \multirow{5}{*}{backgr. knowledge} & \none & \multirow{5}{*}{Y impl.} & healthcare &\multirow{5}{*}{neural network}& \none & \none &\multirow{5}{*}{\cite{DiFrancescomarinoetal:2017}} \\
		& sequence of& & & \none & & automotive & & \none &	\none &	\\ 
		& future activities & & & \none & & financial & & \none &	\none &	\\ 
		& & & & \none & & public admin. & & \none &	\none &	\\ 
		& & & & \none & & customer supp. & & \none &	\none &	\\ \cline{4-11}	
		& & & \none & \none & \multirow{2}{*}{Y} & financial &\multirow{2}{*}{neural network}& \none & \none &\multirow{2}{*}{\cite{Evermann2017}} \\
		& & & \none & \none & & automotive & & \none &	\none &	\\ \cline{3-11}	
		& & event log (with timestamps) & \none & \none & \multirow{3}{*}{ProM plugin} & customer supp. &\multirow{3}{*}{neural network}& \none & \none &\multirow{3}{*}{\cite{Polatoetal:2018}} \\
		& & with data &\none & \none & & public admin. & & \none &	\none &	\\ 
		& & & \none & \none & & financial & & \none &	\none &	\\ \cline{3-11}
		& & event log (with timestamps) & \none & \none & \multirow{3}{*}{Y but unavail.} & \multirow{3}{*}{logistics} &\multirow{3}{*}{clustering}& \multirow{3}{*}{regression} & \none &\multirow{3}{*}{\cite{Folinoetal:2014}} \\
		& & with data &\none & \none & &  & & &	\none &	\\ 
		& & and contextual information &\none & \none & & & & &	\none &	\\ \cline{2-11}	
		& & \multirow{10}{*}{event log (with timestamps)} & \none & \none & \multirow{3}{*}{N} & customer supp. &\multirow{3}{*}{neural network}& \none & \none &\multirow{3}{*}{\cite{Taxetal:2017}} \\
	  & & & \none & \none & & financial & & \none &	\none &	\\ 
		& & & \none & \none & & public admin. & & \none &	\none &	\\ \cline{4-11}	
		& sequence of & & \multirow{5}{*}{backgr. knowledge} & \none & \multirow{5}{*}{Y impl.} & healthcare &\multirow{5}{*}{neural network}& \none & \none &\multirow{5}{*}{\cite{DiFrancescomarinoetal:2017}} \\
		& future activity & & & \none & & automotive & & \none &	\none &	\\ 
		& timestamps & & & \none & & financial & & \none &	\none &	\\ 
		& & & & \none & & public admin. & & \none &	\none &	\\ 
		& & & & \none & & customer supp. & & \none &	\none &	\\ \cline{4-11}	
		& & & \none & \none & \multirow{2}{*}{Y} & financial &\multirow{2}{*}{neural network}& \none & \none &\multirow{2}{*}{\cite{Evermann2017}} \\
		& & & \none & \none & & automotive & & \none &	\none &	\\ \hline	

		\multirow{7}{*}{\rotatebox[origin=c]{90}{\textbf{risk}}} & \multirow{7}{*}{risk} &\multirow{2}{*}{event log (with timestamps)} & \none & \none & Y but unavail. & logistics & clustering & classification & \none &\cite{Cesario2016} \\ \cline{7-11}
		& & & labelling funct. & \none & Camunda pl. & financial & similarity-weight. graph & stat. analysis & \none & \cite{Confortietal:2016}\\ \cline{3-11}
    & & & \multirow{4}{*}{threshold(s)} & \none & \multirow{2}{*}{N} & transport & \multirow{2}{*}{neural network} & \none & \none &  \multirow{2}{*}{\cite{Metzgeretal:2017}}\\ 
		& & event log (with timestamps) & & \none & & logistics &  & \none & \none & \\ \cline{5-11}
	  & & with data & & \none & \multirow{2}{*}{ProM plugin} & financial & \multirow{2}{*}{stochastic Petri net} & \none & \none &  \multirow{2}{*}{\cite{Rogge-Soltietal:2015}}\\ 
		& & & & \none & & logistics &  & \none & \none & \\ \cline{3-11}
		& & & labelling funct. & proc. model & Yawl pugin & no valid. & classification & \none & \none & \cite{DBLP:conf/caise/ConfortiLRA13} \\ \cline{1-11}

		\multirow{9}{*}{\rotatebox[origin=c]{90}{\textbf{inter-case metr.}}} & & event log (with timestamps) & \none & \none & Y but unavail. & logistics & clustering & regression & \none &\cite{Cesario2016} \\ \cline{3-11}
		& & & labelling funct.& \none & ProM plugin & unspec. & regression & \none & \none &\cite{Pikaetal:2016} \\ \cline{4-11}
		& & & \multirow{4}{*}{threshold(s)} & \none & \multirow{3}{*}{N} & transport & \multirow{2}{*}{neural net.} & \none & \none &\multirow{2}{*}{\cite{Metzgeretal:2017}} \\ 
		& inter-case & event log (with timestamps) & & \none &  & logistics & & \none & \none &\\ \cline{7-11}
		& metrics & with data & & \none & & no valid. & classification & regression & time series & \cite{Zeng2008}\\ \cline{5-11}
		& & & & \none & Y but unavail. & unspec. & classification & time series & \none & \cite{Zeng2008}\\ \cline{2-11}
		& \multirow{3}{*}{workload} & event log (with timestamps) & \multirow{3}{*}{labelling funct.}  & \none & \multirow{3}{*}{ProM plugin} &  \multirow{3}{*}{no valid.} &  \multirow{3}{*}{classification} & \none & \none & \multirow{3}{*}{\cite{Castellanos2005}}\\ 
    & & with data & & \none & &  & & \none & \none & \\ 
    & & and contextual information & & \none & &  & & \none & \none & \\ \hline		
		
		\multirow{4}{*}{\rotatebox[origin=c]{90}{\textbf{cost}}} & \multirow{4}{*}{cost} & event log (with timestamps) & \multirow{2}{*}{threshold(s)} & \none & \multirow{2}{*}{N} & transport & \multirow{2}{*}{neural net.} & \none & \none &\multirow{2}{*}{\cite{Metzgeretal:2017}} \\
		 & & with data & & \none & & logistics & & \none & \none & \\ \cline{3-11}
		 & & event log (with timestamps) & \multirow{2}{*}{cost schema} & \none & \multirow{2}{*}{ProM plugin} & \multirow{2}{*}{no valid.} & \multirow{2}{*}{trans. system} & \multirow{2}{*}{stat.analysis} & \none &\multirow{2}{*}{\cite{Wynnetal:2014}} \\
     & & with resources & & \none & & & & & \none & \\
		\bottomrule
		\end{tabular}
		}
		\vspace{0.5cm}
	\caption{Predictive process monitoring framework: \texttt{sequence of outcomes/values}, \texttt{risk}, \texttt{inter-case metrics}, \texttt{cost}}
	\label{tab:framework2}
	\end{adjustwidth}
\end{table}


Table~\ref{tab:framework} and \ref{tab:framework2} reports the devised framework.\footnote{For space limitations, in this article, an abridged version of the framework is presented. The complete version of the framework includes additional data and is available for download at \url{https://docs.google.com/spreadsheets/d/1l1enKhKWx_3KqtnUgggrPl1aoJMhvmy9TF9jAM3snas/edit\#gid=959800788}.} By reading it from left to right, in the first column we find the \textbf{prediction type}. Our review shows that the algorithms can be categorized according to six main types of high level categories of prediction types. The first category, \texttt{time prediction},  encompasses all the different aspects of process execution time such as the \texttt{remaining time} or the \texttt{delay}. Time predictions allow companies to offer better services by providing more precise estimates for delivery/waiting time, or to re-assign resources to deal with cases that are predicted to be delayed. The second main category of identified prediction types is related to \texttt{categorical outcome}(s). Such methods predict the probability of a certain predefined outcome, such as if a case will lead to a disruption, to the violation of a constraint, or whether it will be delayed. 
 Outcome-based predictions allow companies to detect anomalies early on so to pro-actively steer them in the desired direction. Furthermore, outcome-based predictions enable companies to optimize their resources by focusing on cases with higher probability of positive resolution.
The third type of prediction type is related to the \texttt{sequence of next outcomes/ values}. These predictions focus on the probability that future sets of events will occur in the execution of a process instance. Process path predictions of this kind allow for instance service organizations to improve customer satisfaction or detect unexpected termination. With this information at hand, companies can effectively manage cases and gain input for process improvements. The fourth prediction type is \texttt{risk}. When elimination of risks is not feasible, reducing and managing risks becomes important. 
 Risk prediction allow companies to detect and preemptively manage the cases so to avoid undesirable effects of unexpected outcomes. The fifth prediction type pertains the \texttt{inter-case metrics}. 
 The final category is related to \texttt{cost} predictions.
 Predictions on process related costs can be used by companies, in similar manner as the previous prediction types, to optimize the management of process instances. Furthermore, companies can improve their budget monitoring and planning with accurate and real-time data of process-related costs.

The next step (second column) in the framework concerns the \textbf{input data}. According to the predictive process monitoring approach, event logs containing different types of information should be provided as input (e.g., \texttt{event logs (with timestamps)}, \texttt{event logs (with timestamps) with data} in order to apply and exploit the algorith. In some cases, together with the event log, other inputs are required. For instance, in case of the outcome-based predictions, the \texttt{labelling function}, e.g., the specific predicate or category to be predicted, is usually required.  

The framework also considers existence of \textbf{tool support}. If a tool has been developed, it is easier for companies to use, evaluate, and understand the applicability, usefulness and potential benefits of a predictive process monitoring technique. Given that tool support is provided, the framework captures the type such as whether it is a stand-alone application or a plugin of a research framework, e.g., a \texttt{ProM plug-in}.

The quality of the algorithms is highly relevant for companies seeking to adopt and adapt them to their own context. Validating such a quality is hence of the utmost importance for companies. However, validation on logs can take different forms. For instance, it can be achieved by using \texttt{synthetic logs}. Such validations can be considered as ``weaker'' as they do not necessarily mirror the complexity and variability of \texttt{real-life logs}.  Algorithms tested on real-life logs reflect industry logs the best and are therefore, considered as ``stronger''. Furthermore, the suitability of an algorithm is better if validated on logs from the same industry domain as the one of the company seeking to use it. As such, the framework makes note of the \textbf{domain} from which the logs originate. When a domain is specified, it indicates that the algorithm has been tested on a log from that domain. If no domain is specified, the algorithm has not been validated on a real-life log. The type of log together with its domain further support companies in the selection of the most suitable method.

At the heart of each predictive process monitoring method lies the specific algorithm used to implement it. The \textbf{family of algorithm} might matter when assessing the suitability of an approach and as such, it is incorporated in the framework. The specific algorithm is not listed in the framework but rather the foundational technique it is based on, such as \texttt{regression}, \texttt{neural networks}, or \texttt{queuing theory}. 

Following our framework along the above outlined parameters, the most suitable predictive process monitoring method(s) can be identified. For instance, if a company is interested in time predictions, it can focus on methods that offer such predilections (first multi-row in the first leftmost column). In so doing, the company is presented with the different aspects of time that can be predicted. When selecting one of the options (second column), the input data required for such predictive analysis is shown. Following this, available tools are presented and the domain in which it has been validated. Finally, the framework presents the family of algorithm. The last column of the framework references the academic research in which the predictive process monitoring method is introduced. In this way, the framework supports companies to navigate and find the best suited predictive process monitoring method for their business needs and context.

\section{Conclusion}
\label{sec:conclusion}
Predictive process monitoring approaches have been growing quite fast in the last few years. If, on the one hand, such a spread of techniques has provided companies with powerful means for analyzing their business processes and making predictions on their future, on the other hand, it could be difficult for them to navigate in such a complex and unkown domain. By means of a systematic literature review in the predictive process monitoring field, we  provide companies with a framework to guide them in the selection of the technique that best fit their needs. 
In the future we plan to empirically evaluate the proposed framework with real company users in order to assess its usefulness in real contexts.

\bibliographystyle{splncs03}
\bibliography{bibliography}

\end{document}